\documentclass{article}

\usepackage{PRIMEarxiv}

\usepackage[utf8]{inputenc} 
\usepackage[T1]{fontenc}    
\usepackage{hyperref}       
\usepackage{url}            
\usepackage{booktabs}       
\usepackage{amsfonts}       
\usepackage{nicefrac}       
\usepackage{microtype}      
\usepackage{lipsum}
\usepackage{fancyhdr}       
\usepackage{graphicx}       

\usepackage{algpseudocode}
\usepackage{savesym}
\usepackage{bm}
\usepackage{url}
\usepackage{changepage} 
\usepackage{array,multirow,cuted}
\newcolumntype{P}[1]{>{\centering\arraybackslash}p{#1}}
\newcolumntype{M}[1]{>{\centering\arraybackslash}m{#1}}

\usepackage{mathrsfs}
\usepackage{amsmath,amssymb,amsfonts}
\usepackage{textcomp}
\usepackage{xcolor}
\savesymbol{appendices}
\usepackage{appendix}
\restoresymbol{App}{appendices}
\usepackage[section]{placeins}
\usepackage{graphicx}
\usepackage{float}
\usepackage{enumerate}
\usepackage[absolute]{textpos}
\usepackage{times}
\usepackage{epsfig}
\usepackage[euler]{textgreek}
\usepackage{hhline}
\usepackage{lipsum}
\usepackage{mathtools, cuted}
\usepackage{enumitem}

\graphicspath{{media/}}     

\title{Deep Learning based Spatially Dependent Acoustical Properties Recovery
}

\author{
  Ruixian Liu \\
  Department of Electrical and Computer Engineering \\
  University of California, San Diego \\
  La Jolla\\
  \texttt{rul188@ucsd.edu} \\
   \And
  Peter Gerstoft \\
  Scripps Institution of Oceanography \\
  University of California, San Diego \\
  La Jolla\\
  \texttt{pgerstoft@ucsd.edu} \\
}

\begin{document}
\maketitle

\begin{abstract}
The physics-informed neural network (PINN) is capable of recovering partial differential equation (PDE) coefficients that remain constant throughout the spatial domain directly from physical measurements. In this work, we propose a spatially dependent physics-informed neural network (SD-PINN), which enables the recovery of coefficients in spatially-dependent PDEs using a single neural network, eliminating the requirement for domain-specific physical expertise. We apply the SD-PINN to spatially-dependent wave equation coefficients recovery to reveal the spatial distribution of acoustical properties in the inhomogeneous medium. The proposed method exhibits robustness to noise owing to the incorporation of a loss function for the physical constraint that the assumed PDE must be satisfied. For the coefficients recovery of spatially two-dimensional PDEs, we store the PDE coefficients at all locations in the 2D region of interest into a matrix and incorporate the low-rank assumption for such a matrix to recover the coefficients at locations without available measurements. 
\end{abstract}


\section{Introduction}
\label{sec:introduction}
Lots of natural phenomena find their mathematical representation in partial differential equations (PDEs), which are inherently composed of multiple terms and coefficients. A PDE describing the dynamics of field $U$ can be written as
\begin{equation}
\label{eq:pde_general}
{\sf N}[{U}]= a_1 U_x+a_2 U_y +a_3 U_t +a_4U_{tt}+ \dots
\end{equation}
where the partial derivatives $U_x,U_y,U_t,...$ are the PDE terms and the $a_1,a_2,...$ are PDE coefficients. The coefficients are often related to the physical properties of the medium and thus are of great interest in many applications. For example, in mechanical vibrations, the coefficients in the wave equation are related to the elastic properties of the medium \cite{sigalas1994elastic}; in electromagnetics, the coefficients in Maxwell's equations are related to the electrical properties of the medium \cite{assous1993finite}. The spatial variation of the physical properties, like the various elasticities due to the various densities of the medium at different locations, leads to spatially-dependent PDE coefficients (e.g., in \eqref{eq:pde_general}, the coefficients become $a_1(x,y),a_2(x,y)$, etc.). Thus by recovering the spatially-dependent PDE coefficients from observations (i.e., measurements of the dynamical field), we can obtain the spatial distribution of the physical properties of the medium.

The recent developments in computing power have enabled data-driven approaches to identify the PDEs directly from measurements \cite{raissi2019physics,raissi2017physics,raissi2017physics2,brunton2016discovering,rudy2017data,schaeffer2018extracting,long2017pde,long2019pde,brunton2019data,camps2018physics,liu2019wave,zhang2018robust,liu2021pde,xu2019dl,pilar2022physics}. Within these methods, the Physics Informed Neural Network (PINN) \cite{raissi2017physics,raissi2017physics2,raissi2019physics} has garnered considerable scholarly interest due to its notable resilience against measurement noise. Given the type of PDE which delineates the active PDE terms, PINN can learn the representation of the function mapping the spatiotemporal coordinate $(\bm{x}_m,t_j)$ (where for spatially 2D cases $\bm{x}_m$ is a vector) to its measurement $u_{mj}$ by a  fully connected feed-forward neural network (FNN) \cite{eldan2016power}\cite{svozil1997introduction} and recover the PDE coefficients. However, the PINN has limitations when the coefficients for the PDEs are spatially dependent, as it assumes the coefficients are identical across the whole region of interest (ROI).

We propose a Spatially Dependent Physics Informed Neural Network (SD-PINN) which can recover spatially-dependent PDEs using only one neural network, in contrast to the more computational inefficient previous works which use two networks\cite{zhang2020physics,zhang2022physics,kamali2023elasticity} in which one network for solving the PDE and the other for coefficients recovery (e.g., shear modulus\cite{zhang2020physics}, plasma frequency\cite{zhang2022physics} and Lamé parameters\cite{kamali2023elasticity}). 

The SD-PINN also works without requiring domain-specific physical knowledge, and thus is more applicable than prior arts relying on it, e.g., the work which employs the stress-strain relationship \cite{zhang2020physics} and the relation between electron cyclotron frequency and background magnetic field\cite{zhang2022physics}.

Meanwhile, storing the PDE coefficients at all locations in the spatially 2D ROI in a matrix and exploiting the low-rank assumption for this PDE coefficient matrix, the method can recover the coefficients at all locations in the ROI from incomplete measurements which are available at only a part of the ROI. These capabilities, which do not exist in the preliminary version of this work for spatially 1D cases\cite{liu2023sd}, allow the method to offer extensive potential applications in the industry wherever it is needed to recover the physical properties at all locations in the ROI but the sensors can only be placed at a part of the locations and suffer from noise, including but not limited to the material diagnostics and geological survey. 

In this work, we use SD-PINN to recover spatially 2D wave equations with spatially-dependent coefficients to reveal the spatial distribution of acoustical properties for inhomogeneous medium. 

Notations: The 2D or 3D matrices are given in bold capitalized letters, the vectors are in bold lowercase letters, and the scalars are in plain letters. For any variable $\mathbf X$ (or $\bm x$, $x$), its estimation is denoted by $\widehat{\mathbf X}$ (or $\widehat{\bm x}$, $\widehat{x}$). The entry at the $i$th row and $j$th column of matrix $\mathbf X$ is denoted by $\mathbf{X}(i,j)$, and $\mathbf{X}^{\rm T}(i,j)$ denotes the entry at the $i$th row and $j$th column of $\mathbf{X}^{\rm T}$ (the transpose of $\mathbf X$).  $\mathcal{P}_{\Omega}(\mathbf{X})$ denotes the span of matrices vanishing outside a region $\Omega$ so that the $(i,j)$th component of $\mathcal{P}_{\Omega}(\mathbf{X})$ equals to $\mathbf{X}(i,j)$ if $(i,j)\in \Omega$ and zero otherwise. The number of entries within $\Omega$ is denoted by $|\Omega|$.

\section{Theory}
With the type of PDE governing the field of interesting dynamics $U$ in the ROI (with $M$ spatial locations and $T$ time steps) assumed known, we recover the spatially dependent coefficients for each term in the assumed PDE within the ROI.
There are true PDE coefficients at only a few locations in the ROI given, the coefficients at all other locations, which consist the majority of the ROI, are recovered from the measurements of $U$. 

The sign information (non-positive or non-negative) of each coefficient is known from the assumed type of PDE, which is determined by the physical background of the PDE and is the same at all locations.
For example, in the wave equation \cite{buckingham2008transient}
\begin{equation}
U_{tt}+\alpha U_t-c^2\nabla^2 U = 0
\label{eq:wave0}
\end{equation}
the coefficient $-c^2$ for $\nabla^2 U$ (the Laplacian of $U$, i.e., $U_{xx}+U_{yy}$) must be non-positive since $c$ is a real number for the phase speed of the wave, and $\alpha$ which is the factor for attenuation must be non-negative for a system without input energy from external sources.

\begin{figure}
    \centering
    {\includegraphics[width=0.7\linewidth]{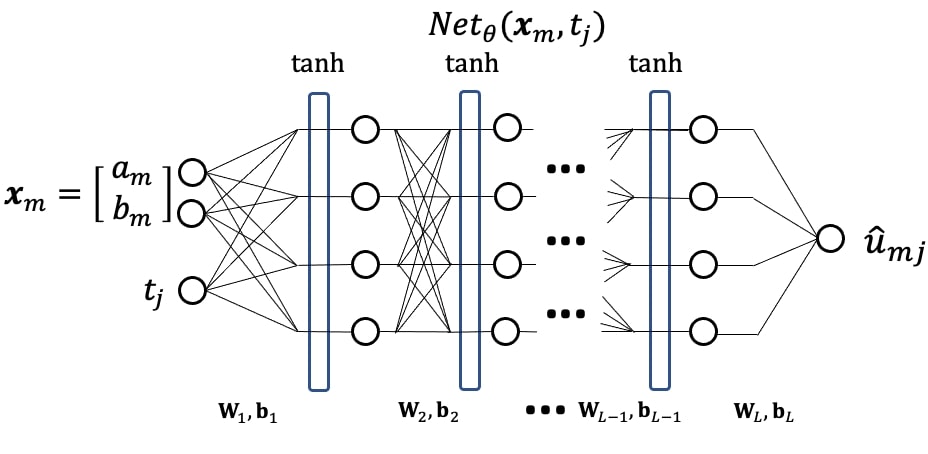}
    }
    \caption{(color online) The FNN used in this work is denoted by a function $\widehat{u}_{mj}=Net_{\theta}(\bm{x}_m,t_j)$. Only one neural network is used for SD-PINN.}
\label{fig:netStruct}
\end{figure}

In an overview of this work, an FNN as Fig.~\ref{fig:netStruct} denoted by a function $Net_\theta$, which is the only neural network used for SD-PINN whose aim is to predict the observation $\widehat{u}_{mj}$ given its coordinates $(\bm{x}_m,t_j)$ is trained. $\theta$ is the parameters (weights and bias) of this FNN. Then PDE terms (i.e., partial derivatives) are computed by automatic differentiation of $Net_\theta$. The spatially-dependent PDE coefficients are then recovered using these partial derivatives computed at various locations. The details are described below.

\subsection{Formulation of spatially-dependent PDEs}
\label{ssec:formulation}
We focus on time-invariant homogeneous PDEs, i.e., there is no source in the ROI and the coefficients do not change with time.

The PDE is written with one term on the left-hand side (LHS) equaling other terms on the right-hand side (RHS). The coefficient of the one term in the LHS is set to one at every location, e.g., for \eqref{eq:wave0},
\begin{equation}
U_{tt}=-\alpha U_t+c^2\nabla^2 U ~.
\label{eq:wave1}
\end{equation}
Our task is to recover the coefficients for all terms in the RHS for all locations.

We denote the LHS at the location $\bm{x}_m$ and time step $t_j$ by $\ell_m^j$. The RHS $r_m^j$ contains $K$ terms $r_{mk}^j$, each of which is a product of a time-invariant coefficient $\lambda_{mk}$ and a PDE term $d^j_{mk}$. So the LHS equaling RHS gives:
\begin{equation}
    \ell_m^j = r_{m}^j = \sum_{k=1}^K r_{mk}^j = \sum_{k=1}^K \lambda_{mk}d^j_{mk}~.
\label{eq:lhs1}
\end{equation}
For example, the wave equation \eqref{eq:wave1} is rewritten as
\begin{equation}
    (U_{tt})_m^j = -\alpha_m (U_t)_m^j+c_m^2(\nabla^2 U)_m^j = \sum_{k=1}^K \lambda_{mk}d^j_{mk}~
\label{eq:waveeqLR}
\end{equation}
where $K=2$, $\lambda_{m1}=-\alpha_m$, $\lambda_{m2}=c_m^2$, $d^j_{m1}=(U_t)_m^j$ and $d^j_{m2}=(\nabla^2 U)_m^j=(U_{xx})_m^j+(U_{yy})_m^j$.
Thus the PDEs at all locations and time steps are written as
\begin{equation}
    \ell_m^j = r_{m}^j=\sum_{k=1}^K \lambda_{mk}d^j_{mk}~,~\forall m,~\forall j~.
\label{eq:waveeq3_sd}
\end{equation}

From \eqref{eq:lhs1}, we can write the RHS for all the locations and PDE terms at time $t_j$ in a matrix as
{
\begin{equation}
\scriptsize
\hspace{-1mm}
\begin{bmatrix}
r^j_{11}& \cdots& r^j_{1K}\\
r^j_{21}& \cdots& r^j_{2K}\\
\vdots & \cdots & \vdots \\
r^j_{M1}& \cdots& r^j_{MK}
\end{bmatrix}
=
\begin{bmatrix}
\lambda_{11}& \cdots & \lambda_{1K}\\
\lambda_{21}& \cdots& \lambda_{2K}\\
\vdots & \cdots & \vdots \\
\lambda_{M1}& \cdots& \lambda_{MK}
\end{bmatrix}
\circ
\begin{bmatrix}
d^j_{11}& \cdots& d^j_{1K}\\
d^j_{21}& \cdots& d^j_{2K}\\
\vdots & \cdots & \vdots \\
d^j_{M1}& \cdots& d^j_{MK}
\end{bmatrix}
\normalsize
\label{eq:lhsmat}
\end{equation}
}where the $MK$ unknown $\lambda_{mk}$ are the coefficients to be recovered and $\circ$ is the Hadamard product. This differs from the conventional PINN, where only a vector of coefficients $[\lambda_1,\dots,\lambda_K]$ is recovered since the PDE is assumed to be spatially independent.
The SD-PINN is demonstrated using the wave equation \eqref{eq:waveeqLR} as an example, but it works the same way for other PDEs. 

\subsection{Low-rank assumption for the spatial variation of coefficients }
\label{ssec:LR}

In this work, we consider spatially 2D cases and assume the ROI to be a rectangular area with $M=M_1M_2$, thus the measurements of the dynamical field acquired at $T$ time steps are stored in a 3D matrix $\mathbf{U}\in\mathbb{R}^{M_1\times M_2\times T}$. By reshaping the $M\times 1$ vector for the coefficients of the $k$th term in \eqref{eq:lhsmat} into an $M_1\times M_2$ matrix and moving the index of the PDE term $k$ to the 3rd dimension, the coefficients will be stored in a 3D matrix containing $K$ slices in $\mathbb{R}^{M_1\times M_2}$ with its $k$th slice being the spatially-dependent coefficients of the $k$th PDE term denoted as: 
\begin{equation}
\boldsymbol{\Lambda}_k=
\begin{bmatrix}
\lambda_{11k}& \cdots& \lambda_{1M_2k}\\
\lambda_{21k}& \cdots& \lambda_{2M_2k}\\
\vdots & \cdots & \vdots \\
\lambda_{M_1 1k}& \cdots& \lambda_{M_1M_2k}~
\end{bmatrix}~.
\label{eq:LAMBDA_k}
\end{equation}
The $\bm{x}_m$ represents the $m$th location in the ROI and is a vector containing a row index and a column index. The objective of SD-PINN is to find an estimation $\widehat{\boldsymbol{\Lambda}}_k$ for all entries of $\boldsymbol{\Lambda}_k$ for all $k$ based on a few given entries.

In real-world scenarios, the spatial variations of the physical properties of the medium for the dynamics are not random, as the properties at a certain point are influenced by the surrounding medium. The decreased degrees of freedom are represented by a lower rank for $\mathbf{\Lambda}_k$, which is smaller than ${\rm{min}}(M_1, M_2)$.

For $\boldsymbol{\Lambda}_k\in\mathbb{R}^{M_1\times M_2}$ with rank $r_k$, there exist two smaller matrices with $r_k$ columns whose multiplication is $\boldsymbol{\Lambda}_k$\cite{chen2018harnessing}:

\begin{equation}
\boldsymbol{\Lambda}_k = \boldsymbol{\mathcal{U}}_k\boldsymbol{\mathcal{V}}_k^{\rm T}
\label{eq:SVDDecomp}
\end{equation}
where $\boldsymbol{\mathcal{U}}_k\in\mathbb{R}^{M_1\times r_k}$ and $\boldsymbol{\mathcal{V}}_k\in\mathbb{R}^{M_2\times r_k}$. 

The method aims to find  $\widehat{\boldsymbol{\mathcal{U}}}_k\in\mathbb{R}^{M_1\times r_k}$ and $\widehat{\boldsymbol{\mathcal V}}_k\in\mathbb{R}^{M_2\times r_k}$ for all $k$ which satisfy $\widehat{\boldsymbol{\mathcal{U}}}_k\widehat{\boldsymbol{\mathcal V}}_k^{\rm T}=\widehat{\boldsymbol{\Lambda}}_k$, such that $\widehat{\boldsymbol{\Lambda}}_k\approx \boldsymbol{\Lambda}_k$ and specifically $\mathcal{P}_{\Omega}(\widehat{\boldsymbol{\Lambda}}_k)=\mathcal{P}_{\Omega}(\boldsymbol{\Lambda}_k)$ by exploiting the information from measurements $\mathbf U$ of the dynamical field governed by PDEs parameterized by $\{\boldsymbol{\Lambda}_k|\forall k=1,\dots,K\}$, where $\Omega$ covers the few locations for the given coefficients. 

Instead of assuming the exact rank of $\boldsymbol{\Lambda}_k$, we assume a reasonable upper limit for that and use it as $r_k$, which not only represents a weaker assumption that is empirically viable but also provides a better recovery as detailed in the following sections. Since the rank of $\boldsymbol{\Lambda}_k$ can be smaller than $r_k$, the column vectors in $\widehat{\boldsymbol{\mathcal{U}}}_k$ and $\widehat{\boldsymbol{\mathcal V}}_k$ are not necessarily linear independent.

By denoting $\boldsymbol{\widehat{\Lambda}}_k=\boldsymbol{\widehat{\mathcal{U}}}_k\boldsymbol{\widehat{\mathcal{V}}}_k^{\rm T}$ we relate the entries in $\boldsymbol{\widehat{\Lambda}}_k$ by the vectors in $\boldsymbol{\widehat{\mathcal{U}}}_k$ and $\boldsymbol{\widehat{\mathcal{V}}}_k$. We thus decrease the number of unknowns to be recovered from $KM_1M_2$ to $\sum_k(M_1+M_2)r_k$ and can use the measurements from only a part of the ROI to recover the properties in the whole ROI. This is valuable when the sensors are insufficient, or there are areas within the ROI where sensors can not be placed.

\subsection{Loss functions}
The used neural network $Net_{\theta}$ parametrized by $\theta$ is an FNN with $L$ layers as shown in Fig.~\ref{fig:netStruct}, whose inputs are the spatial-temporal coordinates $(\bm{x}_m,t_j)$ where $\bm{x}_m=[a_m,b_m]^{\rm T}$ is a vector describing the location indexed by $m$ in the ROI, and outputs are the corresponding estimated measurements $\widehat{u}_{mj}$. 
During the training of the SD-PINN, we minimize the overall loss $loss$ as expressed in Eq.~\eqref{eq:loss2}:
\begin{equation}
loss=loss_{\rm u}+w_{\rm f}\times loss_{\rm f}+w_{\rm g}\times loss_{\rm g}+w_{\rm si}\times loss_{\rm si}~,
\label{eq:loss2}
\end{equation}
which is a linear combination of four individual losses: ${{loss}_{\mathrm{u}}, {loss}_{\mathrm{f}}, {loss}_{\mathrm{g}}, {loss}_{\mathrm{si}}}$ with their weights being $1, w_{\rm f},  w_{\rm g}$ and $w_{\rm si}$ respectively. These losses can be classified into three categories: (i) the data fitting loss $loss_{\mathrm{u}}$ is a function of only the neural network parameters $\theta$ (weights and bias); (ii) the functional loss ${loss}_{\mathrm{f}}$ is a function of both $\theta$ and the PDE coefficients $\lambda$ (which stands for all entries subjected to recovery in $\boldsymbol{\Lambda}_k,~\forall k$); and (iii) the given coefficients loss ${loss}_{\mathrm{g}}$ and sign loss ${loss}_{\mathrm{si}}$ are functions of only the PDE coefficients $\lambda$. 

The $loss$ \eqref{eq:loss2} is minimized via Adam \cite{kingma2014adam}. At the beginning of the network training, all entries in  $\widehat{\boldsymbol{\mathcal{U}}}_k$ and $\widehat{\boldsymbol{\mathcal V}}_k$ for all $k$ are randomly initialized together with $\theta$. The details of these losses are provided below, in which the $\widehat{\boldsymbol{\Lambda}}_k$ is an intermediate variable and during training the gradients are used to update the $\widehat{\boldsymbol{\mathcal{U}}}_k$ and $\widehat{\boldsymbol{\mathcal V}}_k$ essentially. In the optimization related to $\boldsymbol{\widehat{\Lambda}}_k$ (which involves Sec.~\ref{sssec:loss_f},~\ref{sssec:given_coef_loss},~\ref{sssec:sign_loss}), we do not include the  substitution of $\boldsymbol{\widehat{\Lambda}}_k=\boldsymbol{\widehat{\mathcal{U}}}_k\boldsymbol{\widehat{\mathcal{V}}}_k^{\rm T}$ to maintain concise formulaic representation.

\subsubsection{Data fitting loss}
\label{sssec:loss_u}
Given the training samples $\{\bm{x}_m,t_j,u_{mj}\}$ selected from measurements $\mathbf U$, the FNN $Net_{\theta}$ adjusts its parameters (wights and bias) $\theta$ to learn the mapping from coordinates $(\bm{x}_m,t_j)$ to its corresponding measurement $u_{mj}$ by minimizing the $loss_{\rm u}$:
\begin{equation}
loss_{\rm u}(\theta)=\sum_{\bm{x}_m\in{\Omega}_{\rm u}}\sum_{j=1}^T (Net_{\theta}(\bm{x}_m, t_j) - u_{mj})^2~
\label{eq:loss_u_lrsvd}
\end{equation}
where ${\Omega}_{\rm u}$ the set of locations where the measurements are used as training samples to minimize $loss_{\rm u}$.

\subsubsection{Functional loss}
\label{sssec:loss_f}
After $\widehat{u}_{mj}$ is computed by $Net_{\theta}(\bm{x}_m, t_j)$, we compute the PDE terms $\widehat{\ell}^j_{m}$ and $\widehat{d}^j_{mk}$ by automatic differentiation \cite{baydin2018automatic}. For example, the $(U_t)_m^j$ is computed as $\left.\frac{\partial Net_{\theta}(\bm{x},t)}{\partial t}\right\vert_{\bm{x}=\bm{x}_m,t=t_j}$, which is a function of $(\bm{x}, t)$ parametrized by $\theta$. It can also be deemed as a function of $\theta$ parametrized by $\{\bm{x}=\bm{x}_m,t=t_j\}$ if we want to optimize $\theta$ using it, and thus  $\{\widehat{\ell}^j_{m}, \widehat{d}^j_{mk}\}$ can be written as  $\{\widehat{\ell}^j_{m}(\theta), \widehat{d}^j_{mk}(\theta)\}$.

The computation of $\widehat{\ell}^j_{m}$ and $\widehat{d}^j_{mk}$ allows us to introduce $loss_{\rm f}$, by minimizing which we recover the PDE coefficients $\lambda$ and prevent the $Net_{\theta}$ from overfitting the measurements when there is noise in the training samples. The $loss_{\rm f}$ is
\begin{equation}
\begin{aligned}
    loss_{\rm f}(\theta,\lambda)&=\sum_{j\in I_{\rm t}}^{}\sum_{m\in I_{\rm m}}^{}(\widehat{\ell}^j_{m}(\theta) - (\sum_{k=1}^K \widehat{\lambda}_{mk}\widehat{d}^j_{mk}(\theta)))^2\\
    &=\sum_{t\in I_t}\|\widehat{\mathbf{L}}^t(\theta)-\sum_{k}\widehat{\boldsymbol{\Lambda}}_k\circ \widehat{\mathbf{D}}_k^t(\theta)\|_{F}^2
\label{eq:loss_f}
\end{aligned}
\end{equation}
with 
\begin{eqnarray}
    \widehat{\mathbf{L}}^t(\theta)=
    \begin{bmatrix}
        &\widehat{\ell}_{11}^t(\theta)&\cdots &\widehat{\ell}_{1M_2}^t(\theta)\\
        &\vdots &\cdots &\vdots\\
        &\widehat{\ell}_{M_11}^t(\theta) &\cdots &\widehat{\ell}_{M_1M_2}^t(\theta)
    \end{bmatrix},
    \\
    \widehat{\mathbf{D}}_k^t(\theta)=
    \begin{bmatrix}
        &\widehat{d}_{11k}^t(\theta)&\cdots &\widehat{d}_{1M_2k}^t(\theta)\\
        &\vdots &\cdots &\vdots\\
        &\widehat{d}_{M_11k}^t(\theta) &\cdots &\widehat{d}_{M_1M_2k}^t(\theta)
    \end{bmatrix}
\normalsize
\label{eq:LDmat}
\end{eqnarray}
where $I_{\rm m}$ is the set of location indices $m$ 
corresponding to all $\bm{x}_m$ used in $loss_{\rm f}$. As indicated by \eqref{eq:LDmat}, $I_m$ covers all $M_1M_2$ locations within the ROI for our experiments. The $I_{\rm t}$ is the set of time steps used for $loss_{\rm f}$, and is chosen as all time steps from 1 to $T$.



In addition to recovering PDE coefficients $\lambda$, the $loss_{\rm f}$ also benefits the training of neural network parameters $\theta$ by encouraging $Net_{\theta}$ to provide the correct partial derivatives as the PDE terms.
If we only use the $loss_{\rm u}$ to train the network, although we can quickly make the neural network predict the dynamic field itself more accurately, the field's partial derivatives computed by automatic differentiation (AD) are not sufficiently accurate. This is because there are multiple neural network parameters $\theta$ that can make $\widehat{u}_{mj}=Net_{\theta}(\bm{x}_m,t_j)$ approximately equal to the true $u_{mj}$, but for different $\theta$, the AD (for example, $\left.\frac{\partial Net_{\theta}(\bm{x},t)}{\partial t}\right\vert_{\bm{x}=\bm{x}_m,t=t_j}$) are different. 

In addition to recovering the PDE coefficients $\mathbf{\lambda}$, the $loss_{\rm f}$ also encourages $\theta$ to be the one that makes the AD of $Net_{\theta}$ work well as the PDE terms. Without $loss_{\rm f}$, the AD based on the $\theta$ optimized purely on $loss_{\rm u}$ can not simulate the true partial differentiation of ${U}$. 

\subsubsection{Given coefficients loss}
\label{sssec:given_coef_loss}

Let there be $p_{k}$ entries in ${\boldsymbol{\Lambda}}_k$   from a sub-region $\Omega$ of the ROI  known $a$ $priori$, we thus have ${loss}_{\rm{g}}$ (where $\rm{g}$ stands for ``given'') as
\begin{equation}
     loss_{\rm g}(\lambda)=\sum_k \sum_{(a,b)\in\Omega}(\widehat{\boldsymbol{\Lambda}}_k(a,b)- \boldsymbol{\Lambda}_k(a,b))^2~
\label{eq:loss_g_lrsvd}
\end{equation}
where $a$ and $b$ are the row and column indices to enforce all entries within ${\Omega}$ to be identical between the recovered $\widehat{\boldsymbol{\Lambda}}_k$ and true $\boldsymbol{\Lambda}_k$.

\subsubsection{Sign loss}
\label{sssec:sign_loss}
The sign (non-negative or non-positive) for the coefficients in the given type of the PDE is unchanged across the ROI. Thus we can encourage the recovered coefficients to have their assumed signs by minimizing the sign loss 
\begin{equation}
    loss_{\rm{si}}(\lambda)= \sum_{m=1}^M\sum_{k=1}^K{{\rm {ReLU}}(-{\rm{sign}}(\lambda_{mk})\widehat{\lambda}_{mk})}
\label{eq:loss_si}
\end{equation}
where ReLU is the Rectified Linear Unit defined as ${\rm{ReLU}}(x)=x$ for $x>0$ and 0 otherwise, and $\rm{sign}(\lambda_{mk})$ is $1$ for $\lambda_{mk}>0$ or $-1$ for $\lambda_{mk}<0$ depending only on the assumed sign of true $\lambda_{mk}$ and is irrelevant to its approximation $\widehat{\lambda}_{mk}$. Further, the  $\rm{sign}(\lambda_{mk})$ depends only on $k$ because the sign for a given PDE term is assumed the same in the PDE recovered at any location $m$. From \eqref{eq:LAMBDA_k}, the sign loss \eqref{eq:loss_si} is rewritten as
\begin{equation}
    loss_{\rm{si}}(\lambda)=
    \sum_k\sum_{(a,b)\in\rm{ROI}}{{\rm {ReLU}}(-{\rm{sign}}(\boldsymbol{\Lambda}_k(a,b))\widehat{\boldsymbol{\Lambda}}_k(a,b)})
\label{eq:loss_s_lrsvd}
\end{equation}
where the value of ${\rm{sign}}(\boldsymbol{\Lambda}_k(a,b))$ is entirely determined by $k$.

For example, for the wave equation \eqref{eq:wave1} where $\boldsymbol{\Lambda}_1$ denotes $-\alpha$ (non-positive) and $\boldsymbol{\Lambda}_2$ denotes $c^2$ (non-negative), $loss_{\rm {si}}$ is 
\begin{equation}
    loss_{\rm{si}}(\lambda)=
    \sum_{(a,b)\in\rm{ROI}}{{\rm {ReLU}}(\widehat{\boldsymbol{\Lambda}}_1(a,b))}+{\rm {ReLU}}(-\widehat{\boldsymbol{\Lambda}}_2(a,b))~.
\label{eq:loss_s_lrsvd_eg}
\end{equation}
Note that the $\boldsymbol{\Lambda}_k$ and $\widehat{\boldsymbol{\Lambda}}_k$ stand for both the magnitude of the coefficient and its assumed sign. For example, in \eqref{eq:wave1}, the $\boldsymbol{\Lambda}_1$ and $\widehat{\boldsymbol{\Lambda}}_1$ are for $-\alpha$ instead of $\alpha$.

\subsection{Coefficient recovery as a matrix completion problem}
\label{ssec:matrix_completion}

The spatially dependent PDE coefficients recovery can be performed as a matrix completion problem \cite{chen2018harnessing,davenport2016overview,lin2010augmented,bernstein2020typical,klopp2014noisy}. Assuming that for $\boldsymbol{\Lambda}_k$ there are $p_k$ entries known with their spatial locations covered by  ${\Omega}$ (a sub-region of the ROI),
the goal of coefficients recovery is to reconstruct the matrix $\mathbf{\Lambda}_k$ from these known entries subject to the constraint ${\rm{rank}}(\mathbf{\Lambda}_k)\leq r_k$.

We discuss two factors that affect the coefficients recovery: the locations of given coefficients and the number of columns of $\boldsymbol{\widehat{\mathcal{U}}}_k$ and $\boldsymbol{\widehat{\mathcal V}}_k$ (i.e., $r_k$).

\subsubsection{Locations of given coefficients}
\label{sssec:benefits_locs}
For $\boldsymbol{\widehat{\mathcal{U}}}_k\in\mathbb{R}^{M_1\times r_k}$ and $\boldsymbol{\widehat{\mathcal V}}_k\in\mathbb{R}^{M_2\times r_k}$ subjected to recovery, 
the equation $\boldsymbol{\widehat{\mathcal{U}}}_k\boldsymbol{\widehat{\mathcal{V}}}_k^{\rm T}=\boldsymbol{\widehat{\Lambda}}_k$ where $\boldsymbol{\widehat{\Lambda}}_k=\boldsymbol{{\Lambda}}_k$ at $p_k$ specified entries defines a collection of $p_k$ equations with several variables which are a part of the entries in $\boldsymbol{{\widehat{\mathcal{U}}}}_k$ and $\boldsymbol{{\widehat{\mathcal V}}}_k$:
\begin{equation}
\begin{aligned}
\sum_{i=1}^{r_k}\boldsymbol{\widehat{\mathcal{U}}}_k(a_j,i)\boldsymbol{\widehat{\mathcal{V}}}_k^{\rm T}(i,b_j) &=\boldsymbol{\widehat{\Lambda}}_k(a_j,b_j)= \boldsymbol{\Lambda}_k(a_j,b_j),~
\\
&\forall (a_j,b_j) \in \Omega,
~~j=1,\dots, p_k
\end{aligned}
\label{eq:locs_of_unknowns}
\end{equation}
for $\Omega$ with $|\Omega| = p_k$. 

The number of entries of $\boldsymbol{\widehat{\mathcal{U}}}_k$ involved in these equations is $r_k$ times the number of distinct rows covered by ${\Omega}$: for example, when $p_k=2$, in \eqref{eq:locs_of_unknowns}, if $a_1=a_2$, $r_k$ entries of $\boldsymbol{\widehat{\mathcal{U}}}_k$ are involved; otherwise, $2r_k$ entries are involved. Similarly, the number of entries in $\boldsymbol{\widehat{\mathcal V}}_k$ involved is $r_k$ multiplying the number of distinct columns covered by ${\Omega}$. 
Thus, for a fixed number (i.e., $p_k$) of equations, the more distinct rows and columns covered by ${\Omega}$, the more entries of $\boldsymbol{\widehat{\mathcal{U}}}_k$ and $\boldsymbol{\widehat{\mathcal V}}_k$ are affected by
these $p_k$ known coefficients. If the locations in ${\Omega}$ are concentrated in too few distinct rows and columns, the recovery is difficult because the contribution of the known coefficients is constrained within too few entries of $\boldsymbol{\widehat{\mathcal{U}}}_k$ and $\boldsymbol{\widehat{\mathcal V}}_k$.

\subsubsection{Redundant columns of \texorpdfstring{$\widehat{\boldsymbol{\mathcal{U}}}_k$}{TEXT} and \texorpdfstring{$\widehat{\boldsymbol{\mathcal{V}}}_k$}{TEXT}}
\label{sssec:benefits_ranks}
If the specified $r_k$ which is the number of columns in $\widehat{\boldsymbol{\mathcal{U}}}_k$ and $\widehat{\boldsymbol{\mathcal{V}}}_k$ exceeds the true rank of $\mathbf{\Lambda}_k$ (denoted by $r_{k}^0$), this will be an advantage because more degrees of freedom are allowed for the recovery.  This is intuitive because when $\widehat{\boldsymbol{\mathcal{U}}}_k$ and $\widehat{\boldsymbol{\mathcal{V}}}_k$ have $r_k$ columns, their  ranks can be smaller or equal to $r_k$. 
Thus, the potential $\widehat{\boldsymbol{\Lambda}}_k$ generated by $\widehat{\boldsymbol{\mathcal{U}}}_k$ and $\widehat{\boldsymbol{\mathcal{V}}}_k$ with more columns encompasses the $\widehat{\boldsymbol{\Lambda}}_k$ derived from 
$\widehat{\boldsymbol{\mathcal{U}}}_k$ and $\widehat{\boldsymbol{\mathcal{V}}}_k$ with fewer columns. In other words, the
potential $\widehat{\boldsymbol{\Lambda}}_k$ recovered with a higher upper limit of its rank includes those recovered with a lower upper limit, but the reverse is not true.

Meanwhile, the coefficient recovery does not monotonically improve with the increase in the number of columns $r_k$. If $r_k$ is too large, there are so many degrees of freedom for entries in $\widehat{\boldsymbol{\Lambda}}_k$ that the information of recovered entries at locations with available measurements is insufficient to confidently determine the values of entries at other locations.

\begin{figure}[t]
    \centering
    \includegraphics[width=0.5\linewidth]{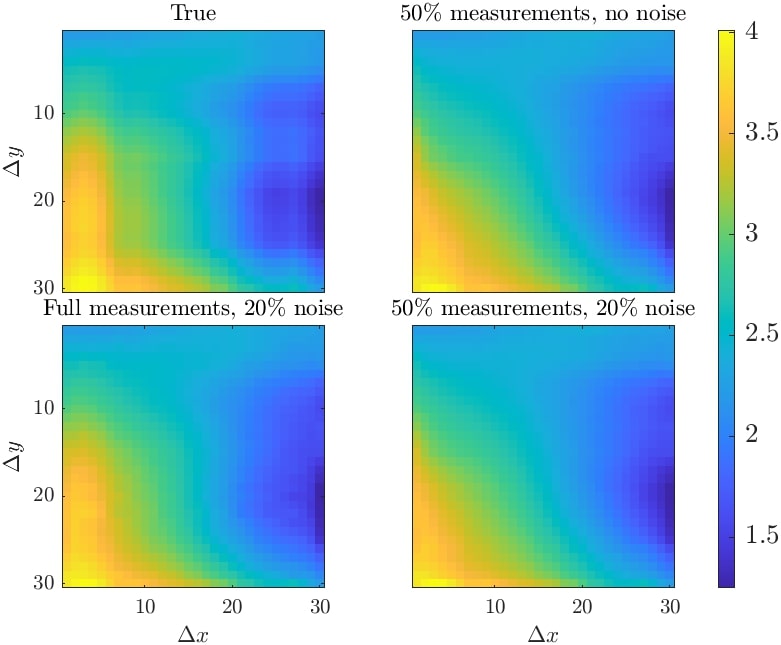}
    \caption{(color online) The true $c^2$ and some recovered $\widehat{c}^2$ in the experiments with $r_1=5$ (at epoch = 4000).}
    \label{fig:rec_c_1}
\end{figure}

\section{Numerical experiments}
\label{sec:exp_obsAtten}
In this section, we explore the PDE coefficients recovery as a matrix completion problem. Various datasets $\mathbf U$ containing the measurements of wave fields governed by the wave equation \eqref{eq:waveeqLR} are used, all of which are in the shape of $\mathbb{R}^{30\times 30\times 198}$. The distances between neighboring coordinates are $\Delta x=\Delta y = 0.1~\rm m$ and $\Delta t=0.01~\rm s$. They may be noise-free or noisy and may be complete (i.e., all entries are given) or incomplete to various extents. 

All available data in $\mathbf U$ with their spatiotemporal coordinates are used to train $Net_{\theta}$ and $\{\boldsymbol{\widehat{\mathcal{U}}}_k,\boldsymbol{\widehat{\mathcal{V}}}_k\}$ are recovered while training. 
Thus, the input is the concatenation of $\bm{x}_m$ and $t_j$ which is a three-component vector
and the output is real number $\widehat{u}_{mj}$ which is used to be compared with the true observation $u_{mj}$. For some experiments we mask the measurements within a portion of spatial locations in $\mathbf U$ to simulate incomplete measurement cases, and in such cases the ``all available data in $\mathbf U$'' refers to a subset of $\mathbf U$.

In all experiments, we use $L=5$ layers for the FNN as shown in Fig.~\ref{fig:netStruct}. The activation function $\tanh$ is applied in the $1{\small \sim} 4$ layers. For the weights $\mathbf{W}_l$ of layer $l$, $\mathbf{W}_l\in\mathbb{R}^{200\times 200}$ for $2\leq l\leq 4$, while $\mathbf{W}_1\in\mathbb{R}^{200\times 3}$ and $\mathbf{W}_5\in\mathbb{R}^{1\times 200}$ accommodate the input and output sizes. All weights are initialized by He initializer \cite{he2015delving}. All biases $\bm b_l$ are in $\mathbb{R}^{200}$ except at the last layer where it is a scalar, and they are initialized as zero. The entries in $\boldsymbol{\widehat{\mathcal{U}}}_k$ and $\boldsymbol{\widehat{\mathcal{V}}}_k$ are initialized as samples drawn from zero-mean Gaussian random distribution with a standard deviation of 0.1. We set $w_{\rm f}=0.1, w_{\rm si}=w_{\rm g}=1$ for \eqref{eq:loss2}. The $\Omega_{\rm u}$ for $loss_{\rm u}$ in \eqref{eq:loss_u_lrsvd} is set to be all locations where measurements are available. As measurements at certain locations may be unavailable, the ${\Omega}_{\rm u}$ is not necessarily the whole ROI. For \eqref{eq:loss_f}, $I_{\rm m}$ is for all the $M$ locations in the ROI, and $I_{\rm t}$ is for all $T$ time steps. 

\begin{table}[t]
    \centering
    \scalebox{1}{
     \begin{tabular}{||c|c|c|c |c||} 
     \hline
      Signals & Noise level & ${\rm{RMSE}}_{c^2}$ ($r_1=3$) & ${\rm{RMSE}}_{c^2}$ ($r_1=5$) \\ [0.5ex] 
     \hline
     All & 0 & 0.140 & 0.128
     \\
     All & 10\% & 0.144 & 0.140
     \\
     All & 20\% & 0.132 & 0.131
     \\
     50\%  & 0 & 0.136 & 0.115 
     \\
      50\%  & 10\% & 0.131 & 0.116 
     \\
     50\% & 20\% & 0.137 & 0.135
     \\
     \hline
     \end{tabular}
    }
     \caption{The root mean square error (RMSE) of recovered $\widehat c$ (at epoch = 4000) for various experiments using the waves without attenuation. The small number in the right bottom corner is the epoch at which the $\widehat{c}^2_m$ is extracted.}
    \label{table:Rmse1new4000}
\end{table}

\begin{figure}[t]
    \centering
    \includegraphics[width=0.7\linewidth]
    {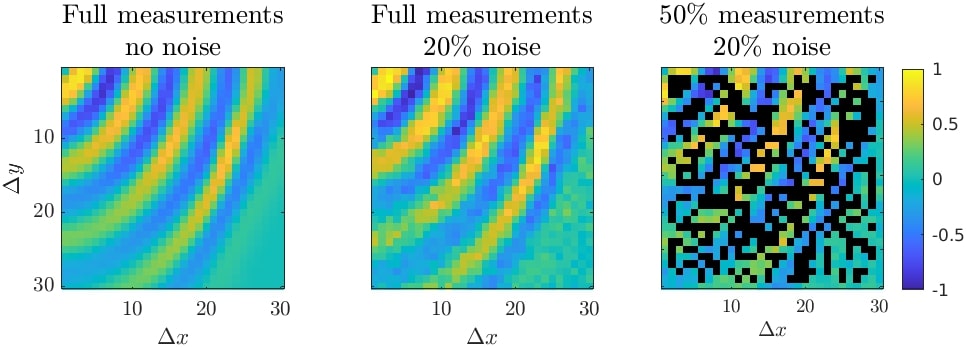}
    \caption{(color online) The clean and noisy signal at frame 100 with full measurements and 50\% measurements. The black pixels denote places without available signals, which are randomly selected.}
    \label{fig:Frames2D}
\end{figure}

\begin{table}[t]
    \centering
    \scalebox{0.9}{
     \begin{tabular}{||c|c|c||} 
     \hline
      Locations of given coefficients & ${\rm{RMSE}}_\alpha$ & ${\rm{RMSE}}_{c^2}$ \\ [0.5ex] 
     \hline
     Diagonal & 1.466 & 0.328
     \\
     Grid & 4.228 & 2.444
     \\
     Random & 1.810  & 0.194
     \\
     \hline
     \end{tabular}
     }
     \caption{The RMSEs between the true and recovered PDE coefficients for different settings of locations for given coefficients.}
    \label{table:RMSEs_locations}
\end{table}

\subsection{Non-attenuating waves}
\label{ssec:non-atten}
We recover the PDEs for non-attenuating waves described in $\mathbf{U}\in\mathbb{R}^{30\times 30\times 198}$ here. The PDE is the spatially-dependent wave equation \eqref{eq:waveeqLR} where $\alpha_m=0$ and $c^2_m$ is distributed as the ``True'' subplot of Figure~\ref{fig:rec_c_1} (rank $= 3$). The unit for the phase speed $c$ is $\rm{m/s}$, and for the attenuation factor $\alpha$ is $\rm s^{-1}$.

In this case, the only coefficient we are recovering is $c^2_m$ for all $m$, and thus $K=1$, the only coefficient matrix $\boldsymbol{\Lambda}_1$ to be recovered in \eqref{eq:LAMBDA_k} is for the phase speeds and has a rank $r_1^0=3$. The set $\Omega$ for all the locations of given coefficients contains four spatial boundaries. The PDE term corresponding to $c^2_m$ is $\nabla^2 U$, which is computed as the sum of $U_{xx}$ and $U_{yy}$, both of which are computed by automatic differentiation of $Net_{\theta}$. We carry out 12 experiments with no noise, 10\% noise, 20\% noise, all measurements available, 50\% measurements available (see Figure \ref{fig:Frames2D}), $r_1=3$ and $r_1=5$ respectively. The noise is additive zero-mean Gaussian noise. The ``10\%" or ``20\% noise'' means the standard deviation (STD) of the Gaussian noise is 10\% or 20\% of the STD of the measurements.
The ``50\% measurements available'' means the available measurements are from all time steps and 50\% spatial locations (randomly selected) of the unknown region in the ROI (i.e., the ROI excluding ${\Omega}$, denoted by $\Omega^c$). 
We measure the recovery results by the root mean square error (RMSE) which is
\begin{equation}
{\rm {RMSE}}_{c^2} = \sqrt{\frac{\sum_{m\in\Omega^c}|\widehat{c}^2_m-c^2_m|^2}{|\Omega^c|}}
\label{eq:rmse_c2}
\end{equation}
where $|\Omega^c|$ is the cardinality of set $\Omega^c$ and summarize them in Table~\ref{table:Rmse1new4000}. The results of the recovery for some experiments are shown in Figure~\ref{fig:rec_c_1}. Both the RMSEs and the graphical demonstrations show that the recovery is satisfactory.

\begin{figure}[t]
\centering
\includegraphics[width=0.6\linewidth]{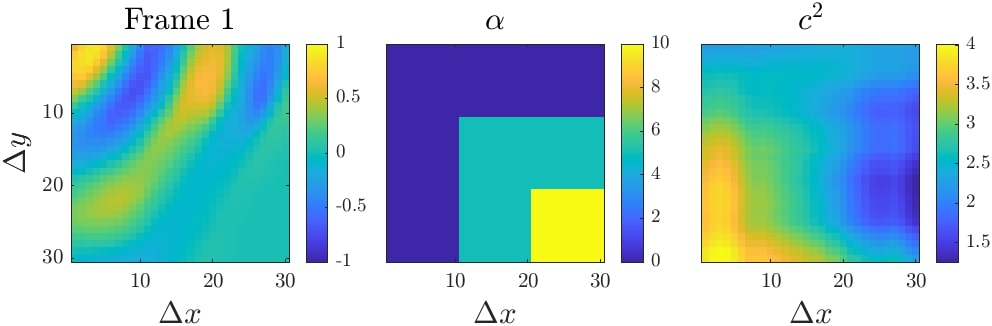}
\caption{(color online) One frame of the wavefield for the attenuating wave with its true $\alpha$ and $c^2$, where $\max (\alpha)=10$.}
\label{fig:Atten10Prop}
\end{figure}

\begin{figure}
\centering
\includegraphics[width=0.7\linewidth]
{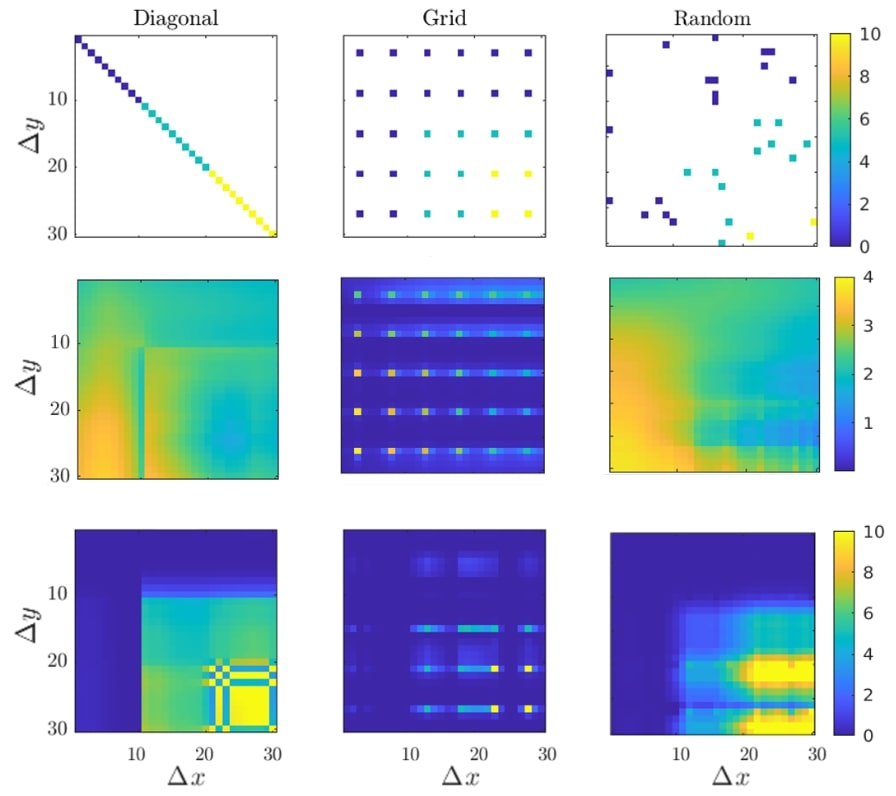}
\caption{(color online) Recovery of PDE coefficients (at 6000th epoch) with 30 entries given: 1st row: the locations of the given entries with the colors representing true $\alpha$ (white pixels are for locations without given coefficients, i.e., $\Omega^c$); 2nd row: recovered $\widehat{c}^2$; 3rd row: recovered $\widehat{\alpha}$.}
\label{fig:Rec_coef_30}
\end{figure}

\begin{figure}
    \centering
    \includegraphics[width=0.35\linewidth]
    {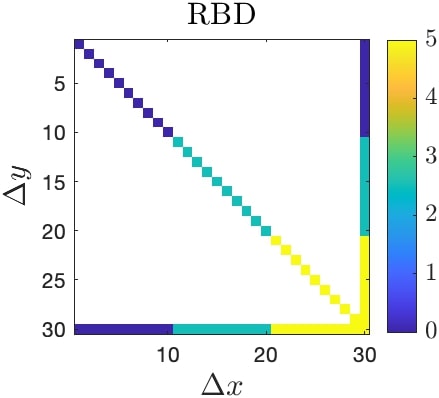}
    \vspace{-3mm}
    \caption{(color online) The locations of the given coefficients which include the right, bottom boundaries, and the diagonal (RBD). There are 88 locations in total. The colors indicate the $\alpha$ at these locations, and white pixels are for locations without given coefficients.}
    \label{fig:RBD_loc}
\end{figure}

\subsection{Attenuating waves}
\label{ssec:atten_waves}
\subsubsection{Locations of given coefficients}
\label{sssec:loc_coefs}
We experiment with a dataset $\mathbf{U}\in\mathbb{R}^{30\times 30\times 198}$ showing an attenuating wavefield. One frame of the field together with its true $\alpha$ and $c^2$ is shown in Fig.~\ref{fig:Atten10Prop}. The spatial variation of $c^2$ is the same as the dataset in Sec.~\ref{ssec:non-atten} so that its rank is 3, and the rank for $\alpha$ is 2. The PDE is the wave equation \eqref{eq:waveeqLR} and thus $K=2$, the $\boldsymbol{\Lambda}_1$ is for $-\alpha$ and $\boldsymbol{\Lambda}_2$ is for $c^2$. The true rank for $\boldsymbol{\Lambda}_1$ is $r_1^0=2$ and for $\boldsymbol{\Lambda}_2$ is $r_2^0=3$. Although $\boldsymbol{\Lambda}_1$ stands for $-\alpha$, we show $\alpha$ in the subsequent figures as it more directly represents the physical properties of the medium.

For this wavefield, we first conduct three experiments with different settings of the locations for the given PDE coefficients. Unlike before, the coefficients on the boundaries are unknown here. In the overall 900 spacial locations within the ROI, the set of locations ${\Omega}$ for given coefficients covers 30 entries, which are on the diagonal, evenly spaced grids, and randomly selected locations respectively. We set $r_1=2,r_2=3$ to run the recovery, the same as true ranks. After 6000 epochs, the coefficients recovery results are summarized in Fig.~\ref{fig:Rec_coef_30}. Compared to Fig.~\ref{fig:Atten10Prop}, it is visibly evident that the recovery of ``diagonal'' is approximately equivalent to ``random'', and both are significantly superior to ``grid''. The RMSEs between the true and recovered coefficients are given in Table~\ref{table:RMSEs_locations}, where ${\rm{RMSE}}_\alpha$ is defined as
\begin{equation}
{\rm {RMSE}}_\alpha = \sqrt{\frac{\sum_{m\in\Omega^c}|\widehat{\alpha}_m-\alpha_m|^2}{|\Omega^c|}}
\label{eq:rmse_alpha}
\end{equation}

The results coincide with our conjecture that when the locations of given coefficients are too concentrated in a few distinct rows and columns, the recovery is hard. For the ``diagonal'', ``random'' and ``grid'', the numbers of distinct rows where the coefficients are given are 30, 20, and 5; and the numbers of distinct columns are 30, 18, and 6 respectively.

\begin{figure}
    \centering
    \includegraphics[width=0.7\linewidth]
    {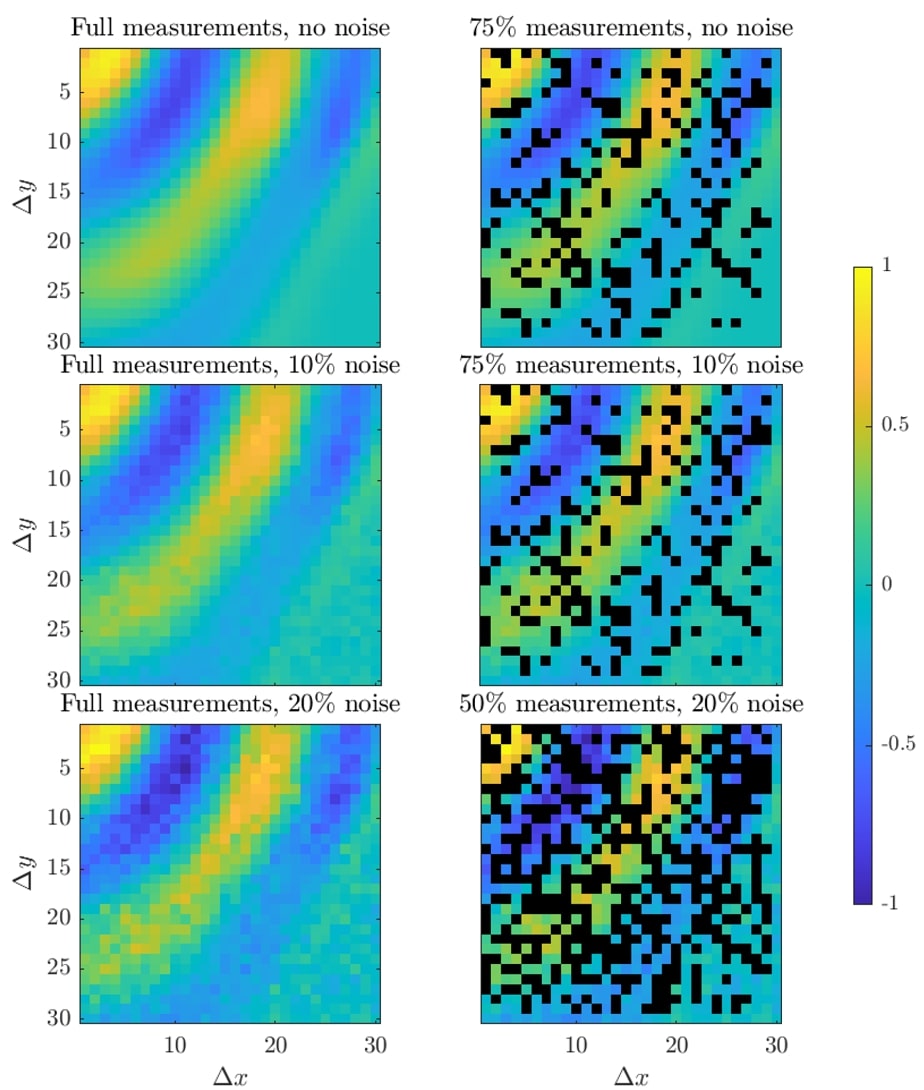}
    \caption{(color online) One frame of the wave field with various percentages of observations and noise levels. The randomly selected black pixels stand for locations without available measurements.}
    \label{fig:FigAttenField5}
\end{figure}

\begin{figure}
\centering
\includegraphics[width=0.7\linewidth]
{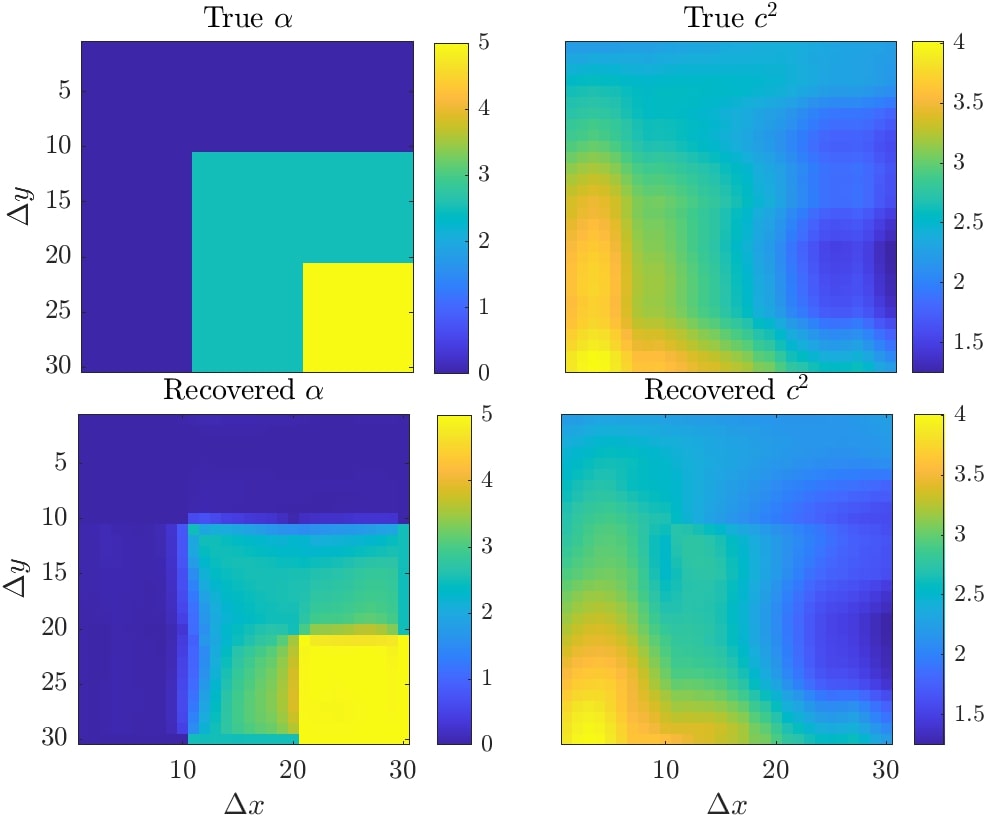}
\caption{(color online) Recovered $\widehat{\alpha}$ and $\widehat{c}^2$ at epoch 5000 given 88 entries on the right, bottom and diagonal for the ground truth ($r_1^0=2$ for $\alpha$, $r_2^0=3$ for $c^2$), using $r_1=r_2=5$ and fully-measured noise-free signals.}
\label{fig:rec_c_a_RBD_r55_epo5000}
\end{figure}

\begin{figure}
\centering
\includegraphics[width=0.7\linewidth]
{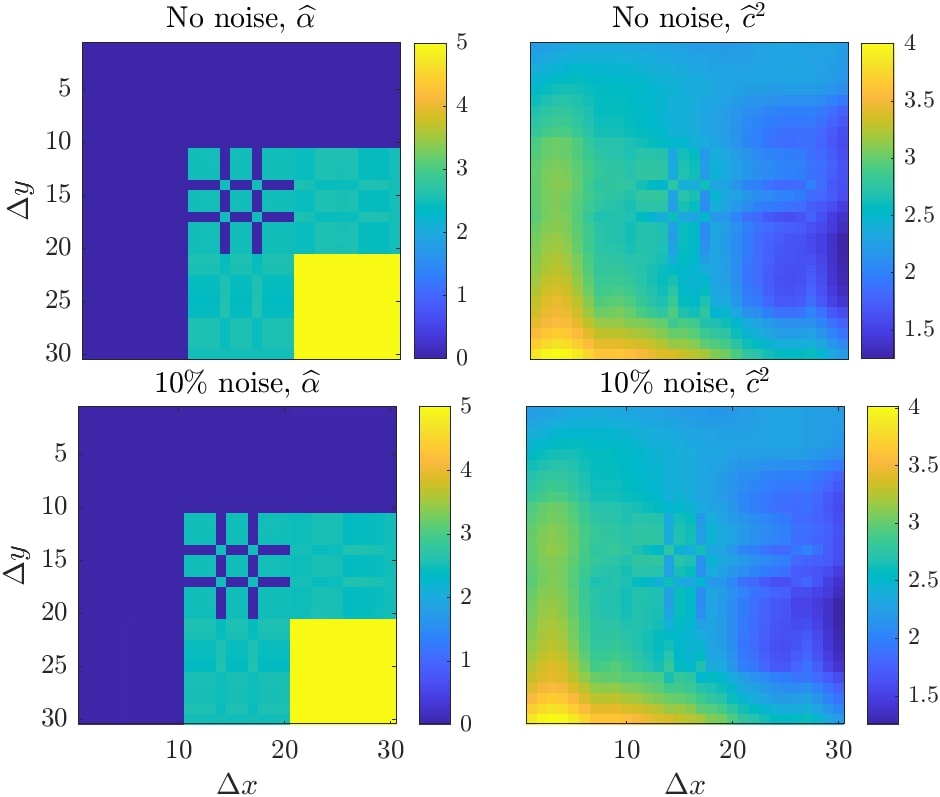}
\caption{(color online) For the 75\%-measured noise-free and noisy signals, the recovered $\widehat{\alpha}$ and $\widehat{c}^2$ at epoch 5000 given 88 entries on the right, bottom and diagonal, using $r_1=2, r_2=3$.}
\label{fig:rec_c_a_RBD_r32_epo5000}
\end{figure}

\subsubsection{Redundant columns of \texorpdfstring{$\widehat{\boldsymbol{\mathcal{U}}}_k$}{TEXT} and \texorpdfstring{$\widehat{\boldsymbol{\mathcal{V}}}_k$}{TEXT}}
\label{sssec:exp_redund_cols}

In this section, we recover the PDE coefficients of an attenuating wavefield with one frame shown in Fig.~\ref{fig:FigAttenField5} and the true coefficients shown in Fig.~\ref{fig:rec_c_a_RBD_r55_epo5000}. For the coefficients, everything is the same as the dataset in Sec.~\ref{sssec:loc_coefs} except that the attenuation is halved. There are $|{\Omega}|=88$ locations of given coefficients on the right boundary + bottom boundary + diagonal (RBD) as shown in Fig.~\ref{fig:RBD_loc}. We carry out 10 experiments where the measurements at all locations are available or at 75\%, 50\% locations are available, and the signal is noise-free or polluted by Gaussian noise whose STD is 10\% or 20\% of the signal's STD. The frame of the signals with various noise levels and availabilities are shown in Fig.~\ref{fig:FigAttenField5}.

\begin{figure}
    \centering
    \includegraphics[width=0.7\linewidth]
    {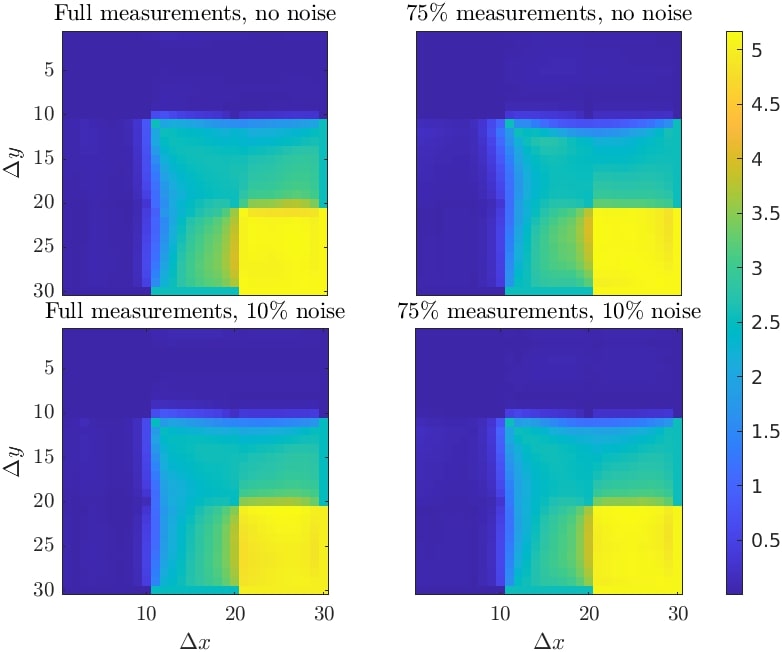}
    \caption{(color online) The recovered $\widehat{\alpha}$ at epoch 5000 for various settings with $r_1=r_2=5$.}
    \label{fig:RecA_Atten5}
\end{figure}

\begin{figure}
\centering
\includegraphics[width=0.7\linewidth]
{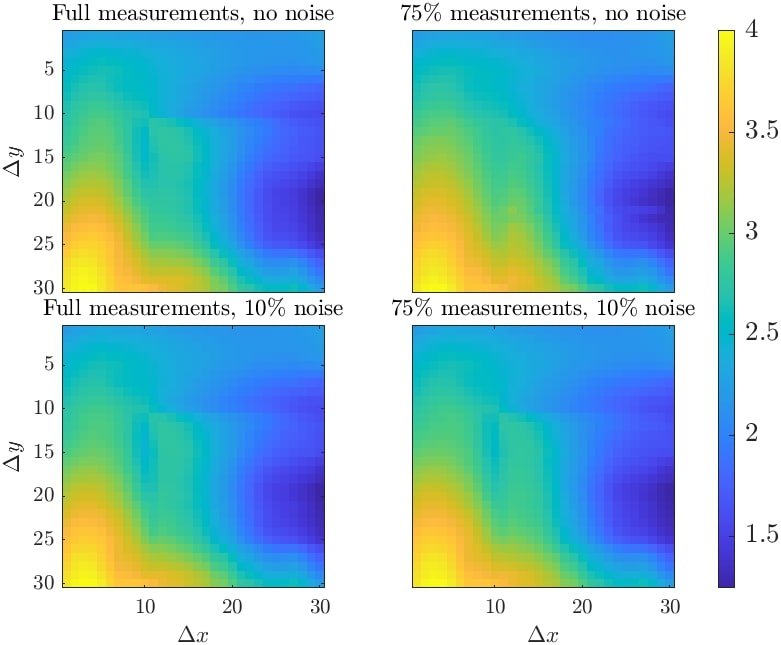}
\caption{(color online) The recovered $\widehat{c}^2$ at epoch  5000 for $r_1=r_2=5$.}
\label{fig:RecC2_Atten5}
\end{figure}

Setting the ranks to be $r_1=r_1^0=2$ and $r_2=r_2^0=3$, the recovery results for 75\% measurements at 5000th epoch are shown in Fig.~\ref{fig:rec_c_a_RBD_r32_epo5000}. Setting the ranks to be $r_1=r_2=5$ which are greater than the true ranks, the recovery results at the 5000th epoch are shown in Fig.~\ref{fig:RecA_Atten5} and \ref{fig:RecC2_Atten5}. Comparing Fig.~\ref{fig:rec_c_a_RBD_r55_epo5000}, Fig.~\ref{fig:rec_c_a_RBD_r32_epo5000}, Fig.~\ref{fig:RecA_Atten5} and \ref{fig:RecC2_Atten5}, allowing additional ranks obviously benefit the coefficients recovery. The RMSEs between the true and recovered coefficients are in Table~\ref{table:RmsePerformA5R5532}.

\begin{figure}
\centering
\includegraphics[width=0.7\linewidth]
{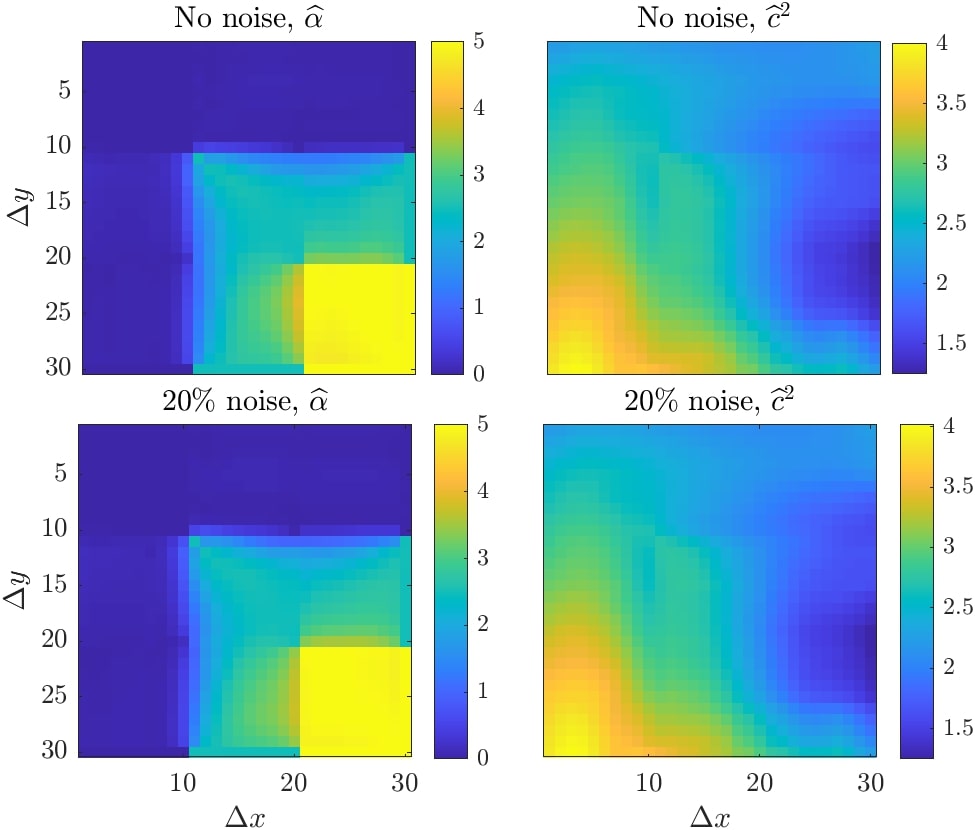}
\caption{(color online) For the 50\%-measured noise-free and noisy signals, the recovered $\widehat{\alpha}$ and $\widehat{c}^2$ at epoch 5000 given 88 entries on the right boundary, bottom boundary, and the diagonal, using $r_1=r_2=5$.}
\label{fig:Atten5_perc50_RBD55_epo5000}
\end{figure}

\begin{figure}
\centering
\vspace{-5mm}
\includegraphics[width=0.7\linewidth]
{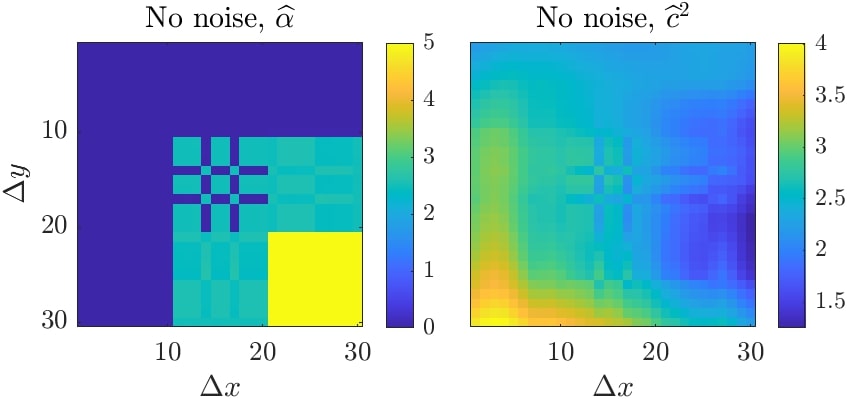}
\caption{(color online) For the 50\%-measured noise-free signals, the recovered $\widehat{\alpha}$ (left) and $\widehat{c}^2$ (right) at epoch 5000 given 88 entries on the right boundary, bottom boundary, and the diagonal, using $r_1=2, r_2=3$.}
\label{fig:rec_c_a_RBD_perc50_r32_epo5000}
\end{figure}

From Table~\ref{table:RmsePerformA5R5532}, we see that the recovery using $r_1=r_2=5$ is better than using $r_1=2,r_2=3$. From the table and Fig.~\ref{fig:Atten5_perc50_RBD55_epo5000}, the recovery using $r_1=r_2=5$ is satisfactory even for the case with noisy data (noise STD = 20\% of signal STD) and 50\% measurements. But when $r_1=2$ and $r_2=3$, the recovery is problematic for the 50\% measurements case as indicated in Fig.~\ref{fig:rec_c_a_RBD_perc50_r32_epo5000}.

\begin{table}
    \centering
    \scalebox{0.83}{
     \begin{tabular}{||c|c|c|c|c||} 
     \hline
      \begin{tabular}{@{}c@{}}Experimental \\ settings\end{tabular} & $r_1$ & $r_2$ & ${\rm{RMSE}}_\alpha$ & ${\rm{RMSE}}_{c^2}$ \\ [0.5ex] 
     \hline
     \begin{tabular}{@{}c@{}}Full measurements \\ no noise\end{tabular} & 5 & 5 & 0.366 & 0.145
     \\
     \hline
     \begin{tabular}{@{}c@{}}75\% measurements \\ no noise\end{tabular} & 5 & 5 & 0.369 & 0.125
     \\
     \hline
     \begin{tabular}{@{}c@{}}50\% measurements \\ no noise\end{tabular} & 5 & 5 & 0.371 & 0.140
     \\
     \hline
     \begin{tabular}{@{}c@{}}50\% measurements \\ no noise\end{tabular} & 2 & 3 & 0.495 & 0.186
    \\
    \hline
     \begin{tabular}{@{}c@{}}Full measurements \\ 10\% noise\end{tabular} & 5 & 5 & 0.359  & 0.153 
     \\
     \hline
     \begin{tabular}{@{}c@{}}75\% measurements \\ 10\% noise\end{tabular} & 5 & 5 & 0.356  & 0.150
     \\
     \hline
     \begin{tabular}{@{}c@{}}75\% measurements \\ no noise\end{tabular} & 2 & 3 & 0.496 & 0.192
     \\
     \hline
     \begin{tabular}{@{}c@{}}75\% measurements \\ 10\% noise\end{tabular} & 2 & 3 & 0.495 & 0.198
     \\
     \hline
      \begin{tabular}{@{}c@{}}Full measurements \\ 20\% noise\end{tabular} & 5 & 5 & 0.400 & 0.134
     \\
     \hline
      \begin{tabular}{@{}c@{}}50\% measurements \\ 20\% noise\end{tabular} & 5 & 5 & 0.398 & 0.139
     \\
     \hline
     \end{tabular}
     }
\caption{RMSEs between true and recovered PDE coefficients for different settings of ranks, signal availability, and noise conditions.}
\label{table:RmsePerformA5R5532}
\end{table}

\section{Comparison with two baseline methods}
\label{sec:comparison}
We compare the coefficient recovery result between the SD-PINN and two baseline methods. Given the noise-free and 50\% measurements of the field in Sec.~\ref{sssec:exp_redund_cols} with the measurements sampled at the same 50\% locations as in Fig.~\ref{fig:FigAttenField5}, the coefficients within all the ROI are recovered by:
(baseline-1) first interpolating the measurements by spline interpolation \cite{akima1970new} to obtain full measurements, and then recover the coefficients for every location iteratively based on the interpolated signals using regression on a dictionary of PDE terms\cite{liu2022data}\cite{liu2023recovery}, which is a simplified version of SINDy\cite{brunton2016discovering} because the dictionary only contains correct PDE terms without redundant ones; 
(baseline-2) first recover the coefficients at a few locations with sufficient measurements, and then use the matrix completion approach \cite{cai2010singular} to recover the coefficients at other locations.

Before diving into the baseline methods, we outline the PDE coefficients recovery by finite difference (FD) \cite{ames2014numerical} with ordinary least squares regression (OLS) which is used in both baseline methods and its limitation. Given the measurements at three consecutive locations along $x$-axis centered at $i$ with $y$-coordinate $j$ and time step $k$: $\{\mathbf{U}(i-1,j,k),\mathbf{U}(i,j,k),\mathbf{U}(i+1,j,k)\}$, the first order spatial derivative along $x$ at $(i,j)$ is computed as $[\mathbf{U}(i+1,j,k)-\mathbf{U}(i-1,j,k)]/{2\Delta x}$ and the 2nd order derivative is $[\mathbf{U}(i+1,j,k)-2\mathbf{U}(i,j,k)+\mathbf{U}(i-1,j,k)]/{\Delta x^2}$. Such calculations can be repeated at all time steps, and thus for location $(i,j)$, we can obtain vectors containing the numerical partial derivatives along $x$-axis as
\begin{equation}
\begin{aligned}
    \mathbf{u}_x^{(i,j)} &= \frac{\mathbf{U}(i+1,j,:)-\mathbf{U}(i-1,j,:)}{2\Delta x}\\
    \mathbf{u}_{xx}^{(i,j)} &= \frac{\mathbf{U}(i+1,j,:)-2\mathbf{U}(i,j,:)+\mathbf{U}(i-1,j,:)}{\Delta x^2}~.
\end{aligned}
\label{eq:ux_uxx}
\end{equation}
The partial derivatives at $(i,j)$ along the $y$-axis are computed similarly. For the partial derivatives along time, only the measurements at $(i,j)$ are sufficient:
\begin{equation}
\begin{aligned}
\mathbf{u}_t^{(i,j)}(k) &= \frac{\mathbf{U}(i,j,k+1)-\mathbf{U}(i,j,k-1)}{2\Delta t}\\
\mathbf{u}_{tt}^{(i,j)}(k) &= \frac{\mathbf{U}(i,j,k+1)-2\mathbf{U}(i,j,k)+\mathbf{U}(i,j,k-1)}{\Delta t^2}
\end{aligned}
\label{eq:ut_utt}
\end{equation}
where $2\leq k\leq \rm{number~of ~time ~steps-1}$.

From \eqref{eq:ux_uxx} and \eqref{eq:ut_utt}, we do not consider the FD evaluated at the boundaries of $\mathbf{U}$ which is defined differently and subjected to larger errors. For the considered dataset $\mathbf{U}\in\mathbb{R}^{30\times 30\times 198}$ in Sec.~\ref{sssec:exp_redund_cols}, $\{\mathbf{u}_x,\mathbf{u}_{xx},\mathbf{u}_y,\mathbf{u}_{yy}\}$ all of which are in $\mathbb{R}^{198}$ can be computed at all locations except the spatial boundaries, so there are $28^2=784$ locations in total. The $\mathbf{u}_t$ and $\mathbf{u}_{tt}$ are also computed at these locations, and according to \eqref{eq:ut_utt}, $\{\mathbf{u}_t^{(i,j)}(k),\mathbf{u}_{tt}^{(i,j)}(k)\}$ are well-defined for $2\leq k\leq 197$. So for each $(i,j)$, we drop first and last entries of $\{\mathbf{u}_t^{(i,j)},\mathbf{u}_{tt}^{(i,j)}\}$ to make them in $\mathbb{R}^{196}$. Similarly, the first and last entries of the spatial derivative vectors are also dropped. Then for $(i,j)$ we construct a matrix as
\begin{equation}
    \boldsymbol{\Phi}^{(i,j)}=[-\mathbf{u}_{t}^{(i,j)},\mathbf{u}_{xx}^{(i,j)}+\mathbf{u}_{yy}^{(i,j)}]\in\mathbb{R}^{196\times 2}
\label{eq:phi}
\end{equation}
and then the coefficients at $(i,j)$ are recovered by OLS as
\begin{equation}
    [\widehat{\alpha}(i,j),~\widehat{c}(i,j)^2]^{\rm T}=\boldsymbol{\Phi}^{(i,j)\dagger}\mathbf{u}_{tt}^{(i,j)}
\label{eq:lsq}
\end{equation}
where $\dagger$ denotes pseudo-inverse.

From the above discussion, the limitation of the FD+OLS method is that for location $(i,j)$, the measurements at all its neighbors $\{(i-1,j),(i+1,j),(i,j-1),(i,j+1)\}$ must exist. This is not true when there are many sensors out of work, e.g., the 50\% measurements as in Fig.~\ref{fig:FigAttenField5}. To recover all coefficients using only partial observations, the two methods are detailed below.

\begin{figure}
\centering
\includegraphics[width=0.7\linewidth]
{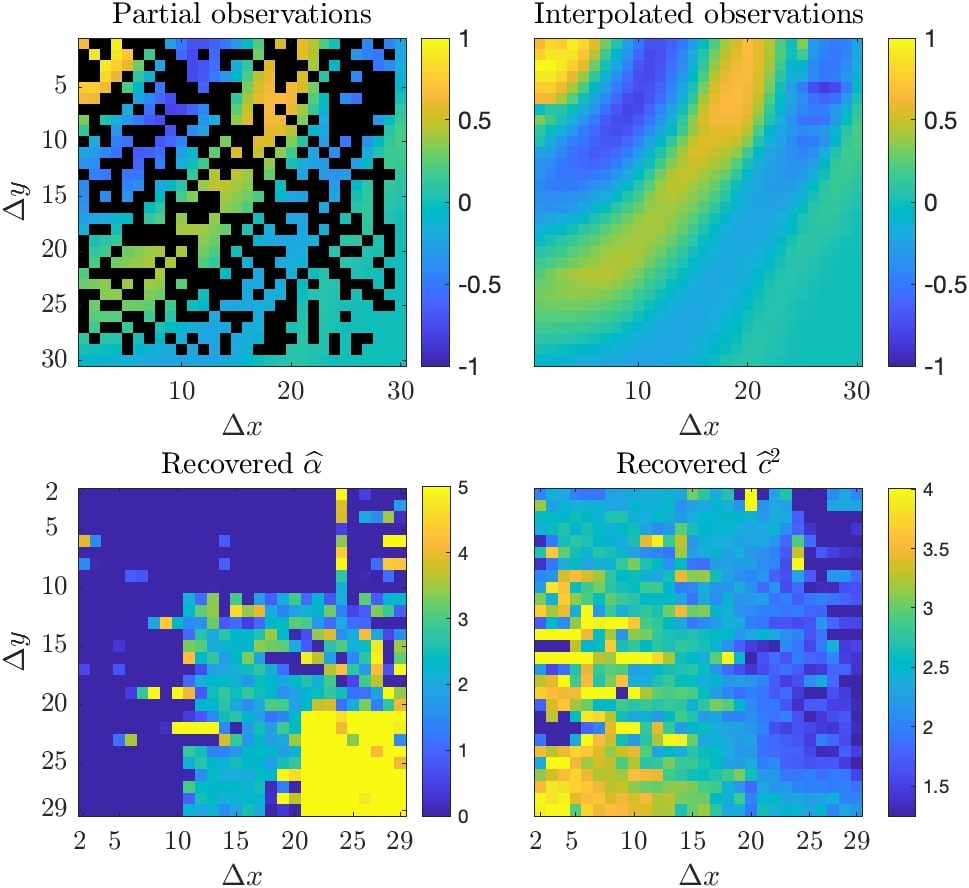}
\caption{(color online) One frame of the 50\% sampled noise-free signal and its interpolation result, and the recovered PDE coefficients via baseline-1. For this method, the recovered coefficients are located within [2,29] for both axes.}
\label{fig:CoefRecoveryBaseline1}
\end{figure}

Details of the baseline-1 method: (1) For every frame of the 50\% measurements, first do spline interpolation row by row, and then do the interpolation again column by column, and in the end average these two interpolation results to be the interpolated signals at this frame. (2) Except for the four boundaries, for each of the $28^2$ locations in the ROI, use the above-mentioned FD+OLS method to recover the coefficients. One frame of the noise-free 50\% measurements with its interpolation result and the recovered coefficients are shown in Fig.~\ref{fig:CoefRecoveryBaseline1}.

\begin{figure}
\centering
\includegraphics[width=0.7\linewidth]
{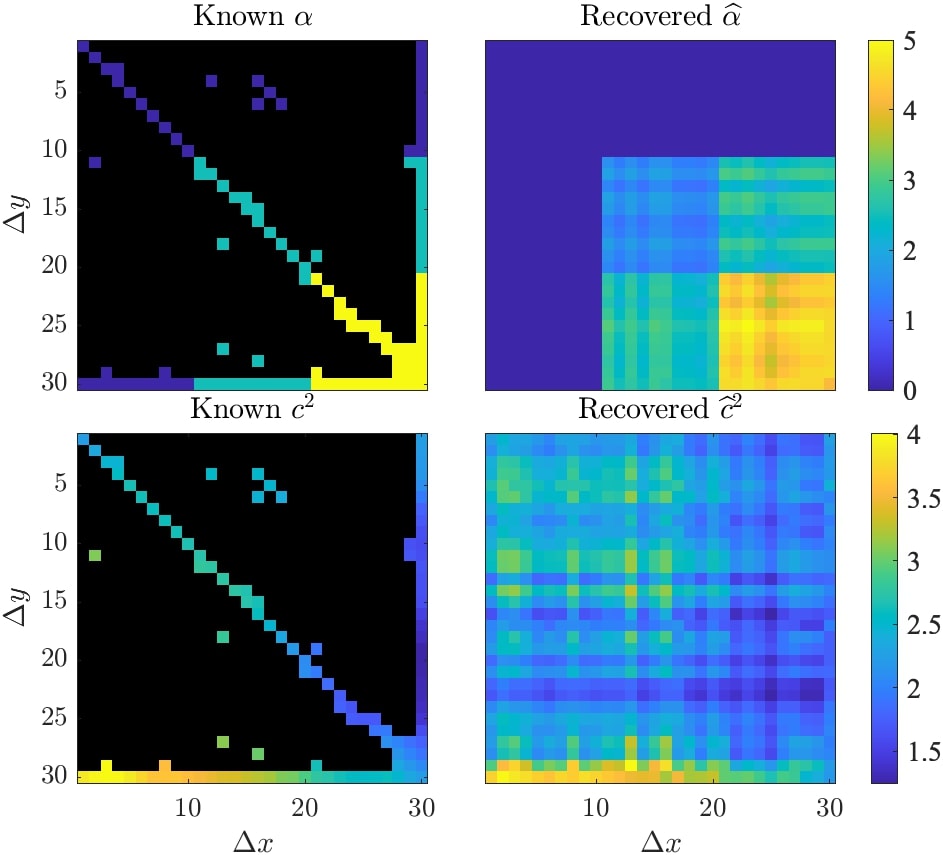}
\caption{(color online) Known and recovered PDE coefficients for the baseline-2 method. The known coefficients include the given ones on the right boundary, bottom boundary, and the diagonal as well as the recovered ones via FD+OLS at a few eligible locations. Black pixels are for locations without known coefficients.}
\label{fig:CoefRecoveryBaseline2}
\end{figure}

Details of the baseline-2 method: (1) In addition to the known coefficients on the bottom boundary, right boundary, and the diagonal, as indicated in Fig.~\ref{fig:FigAttenField5}, there are a few other locations eligible for the spatial derivatives to be computed by FD (for such a location, the measurements are available at itself and all its top, bottom, left and right neighbors). We first recover the coefficients by FD+OLS at these locations. (2) Based on the coefficients that are given and recovered at the few locations mentioned above, we recover the coefficients at other locations via matrix completion by nuclear norm minimization (NNM) \cite{cai2010singular}:
\begin{equation}
\widehat{\boldsymbol{\Lambda}}_k=\arg\min_{{\mathbf X}} \tau\|\mathbf X\|_*+\frac{1}{2}\|\mathbf X\|_F^2~~ 
\text { s.t. } \mathcal{P}_{\Omega}(\mathbf X)=\mathcal{P}_{\Omega}(\boldsymbol{\Lambda}_k) \\
\label{eq:nnm_theory}
\end{equation}
where the nuclear norm $\|\mathbf X\|_*$ is the sum of its singular values and $\|\mathbf X\|_F$ the Frobenius norm. The \eqref{eq:nnm_theory} is solved iteratively from $\mathbf{Y}^0=\mathbf{0}\in\mathbb{R}^{M_1\times M_2}$ with step $\delta$ by
\begin{equation}
\left\{\begin{array}{l}
\mathbf{X}^i=\mathcal{D}_\tau\left(\mathbf{Y}^{i-1}\right) \\
\mathbf{Y}^i=\mathbf{Y}^{i-1}+\delta \mathcal{P}_{\Omega}\left(\boldsymbol{\Lambda}_k-\mathbf{X}^i\right)
\end{array}\right.
\label{eq:nnm_iter}
\end{equation}
where $\mathcal{D}_\tau$ the singular value shrinkage operator, i.e., suppose the  singular value decomposition (SVD) \cite{wall2003singular} of $\mathbf{Y}$ with rank $r$ is
\begin{equation}
\mathbf{Y}=\mathbf{U} \boldsymbol{\Sigma} \mathbf{V}^{\rm T}, \quad \boldsymbol{\Sigma}=\operatorname{diag}\left(\left\{\sigma_i\right\}_{1 \leq i \leq r}\right),
\end{equation}
then
\begin{equation}
\mathcal{D}_\tau(\mathbf{Y}):=\mathbf{U} \mathcal{D}_\tau(\boldsymbol{\Sigma}) \mathbf{V}^{\rm T}, \quad \mathcal{D}_\tau(\boldsymbol{\Sigma})=\operatorname{diag}\left(\left\{\sigma_i-\tau\right\}_{+}\right)
\end{equation}
with $\{t\}_+={\rm{max}}(0,t)$. From \eqref{eq:nnm_theory}, the rank $r_k$ of recovered $\widehat{\boldsymbol{\Lambda}}_k$ is adjustable: as $\tau$ increases, $r_k$ decreases in general. Multiple experiments are carried out using various $\tau$, but none of them provide satisfactory coefficient recovery. Among them, the recovered $\widehat{\boldsymbol{\Lambda}}_1$ with rank 2 (for $\widehat{\alpha}$) and $\widehat{\boldsymbol{\Lambda}}_2$ with rank 3 (for $\widehat{c}^2$) 
together with the known $\alpha$ and $c^2$ on which the recovery is based (including given coefficients and the recovered coefficients via FD+OLS) are shown in Fig.~\ref{fig:CoefRecoveryBaseline2}.

Visual examinations of Fig.~\ref{fig:CoefRecoveryBaseline1} and \ref{fig:CoefRecoveryBaseline2} suggest that the PDE coefficients recovery by the two baseline methods is far poorer than SD-PINN, as shown in Fig.~\ref{fig:Atten5_perc50_RBD55_epo5000}. The RMSEs for the two baseline methods are in Table \ref{table:RmseBaselines}. Compared to Table \ref{table:RmsePerformA5R5532}, except for the recovery of $\widehat{\alpha}$ by baseline-2 which is slightly worse than SD-PINN ($r_1=r_2=5$ case), all other recoveries are much worse than SD-PINN.

\begin{table}[t]
    \centering
    \scalebox{1}{
     \begin{tabular}{||c|c|c||} 
     \hline
      Method & ${\rm{RMSE}}_\alpha$ & ${\rm{RMSE}}_{c^2}$ \\ [0.5ex] 
     \hline
     Baseline-1  & 1.379 & 0.929
     \\
     Baseline-2  & 0.381 & 0.810
     \\
     SD-PINN & 0.371 & 0.140\\
     \hline
     \end{tabular}
     }
\caption{RMSEs between the true and recovered PDE coefficients by two baseline methods from noise-free 50\% measurements sampled at locations indicated in Fig.~\ref{fig:FigAttenField5}, the corresponding SD-PINN result with $r_1=r_2=5$ from Table \ref{table:RmsePerformA5R5532} is included for comparison.}
\label{table:RmseBaselines}
\end{table}

\section{Conclusion}
\label{sec:conclusion}
We propose a spatially-dependent physics-informed neural network (SD-PINN) method to recover the spatially-dependent PDE coefficients from the observations. The PDE coefficients are recovered as the entries of the matrices and the recovery is formulated as a matrix completion problem with low rank constraints which is solved by a neural network. The experiments show that the proposed method can successfully recover the spatially dependent coefficients for the wave equation, and thus can recover the spatial distribution of the acoustical properties including phase speeds and attenuations. The recovery is robust to noise and poor availability of measurements. Its performance is better when the locations of given coefficients are not constrained to too few distinct rows and columns, and is affected by the assumed ranks of the coefficient matrices.

\end{document}